\documentclass{article}
\usepackage[preprint,nonatbib]{neurips_2021}
\usepackage[utf8]{inputenc} 
\usepackage[T1]{fontenc}    
\usepackage{hyperref}       
\usepackage{url}            
\usepackage{booktabs}       
\usepackage{amsfonts}       
\usepackage{nicefrac}       
\usepackage{microtype}      
\usepackage{graphicx}
\usepackage{amsthm}
\usepackage{booktabs}
\usepackage{algorithm}
\usepackage{algorithmic}
\usepackage{amsfonts}
\usepackage{amsmath, bm}
\usepackage{multirow}
\usepackage{bbm}
\usepackage{color}
\usepackage{xcolor}
\usepackage{caption}
\usepackage{subcaption}
\usepackage[toc,page]{appendix}
\usepackage{hyperref}
\usepackage{array}
\usepackage{xhfill}
\usepackage{tabularx}

\usepackage{mathtools,tikz,caption}
\newcolumntype{L}{>{\centering\arraybackslash}m{3cm}}

\newtheorem{definition}{Definition}

\newtheorem{result}{Result}

\DeclareRobustCommand\sampleline[1]{%
  \tikz\draw[#1] (0,0) (0,\the\dimexpr\fontdimen22\textfont2\relax)
  -- (1.5em,\the\dimexpr\fontdimen22\textfont2\relax);%
}

\title{Fair Regression under Sample Selection Bias}
\author{%
  Wei Du \\
  University of Arkansas\\
  Fayetteville, AR 72701 \\
  \texttt{wd005@uark.edu} \\
  \And
  Xintao Wu \\
  University of Arkansas\\
  Fayetteville, AR 72701 \\
  \texttt{xintaowu@uark.edu} \\
   \And
  Hanghang Tong \\
  University of Illinois
  \\
  Urbana, IL 61801 \\
  \texttt{htong@illinois.edu} \\
}

\begin{document}

\maketitle
\begin{abstract}
Recent research on fair regression focused on developing new fairness notions and approximation methods as target variables and even the sensitive attribute are continuous in the regression setting. However, all previous fair regression research assumed the training data and testing data are drawn from the same distributions. This assumption is often violated in real world due to the sample selection bias between the training and testing data. In this paper, we develop a framework for fair regression under sample selection bias when dependent variable values of a set of samples from the training data are missing as a result of another hidden process. Our framework adopts the classic Heckman model for bias correction and the Lagrange duality to achieve fairness in regression based on a variety of fairness notions. Heckman model describes the sample selection process and uses a derived variable called the Inverse Mills Ratio (IMR) to correct sample selection bias. We use fairness inequality and equality constraints to describe a variety of fairness notions and apply the Lagrange duality theory to transform the primal problem into the dual convex optimization.  For the two popular fairness notions, mean difference and mean squared error difference, we derive explicit formulas without iterative optimization, and for Pearson correlation, we derive its conditions of achieving strong duality. We conduct experiments on three real-world datasets and the experimental results demonstrate the approach's effectiveness in terms of both utility and fairness metrics.
\end{abstract}

\section{Introduction}

Fairness has been an increasingly important topic in machine learning. Fair machine learning models aim to learn a function $f$ for a target variable $Y$ using input features $X$ and a sensitive attribute $A$ (e.g., gender), while ensuring the predicted value $\hat{Y}$ fair with respect to $A$ based on some given fairness criterion. Fair machine learning models can be categorized into pre-processing (modifying training data or learning a new representation such that the information correlated to the sensitive attribute is removed), in-processing (adding fairness penalty to the objective function during training), and post-processing (applying perturbation or transformation to model output to reduce prediction unfairness). Much of existing works has focused on classification. In this paper, we focus on fair regression where the target $Y$ is continuous.

Fair regression can be naturally defined as the task of minimizing the expected loss of real-valued predictions, subject to some fairness constraints. Fairness notions under the regression setting are in principle based on some forms of independence, e.g., the independence of model prediction $\hat{Y}$ and sensitive attribute $A$, the independence of prediction error $\hat{Y}-Y$ and sensitive attribute $A$, and the conditional independence of $\hat{Y}$ and $A$ given $Y$. Different from the classification setting,  variables of $Y$ and $\hat{Y}$ (even $A$) become continuous in the regression setting, which requires new fairness notions and constrained optimization techniques.
Researchers have developed quantitative metrics based on moment constraints, such as mean difference \cite{calders2013controlling}, mean squared error difference \cite{johnson2016impartial}, and Pearson correlation \cite{komiyama2018nonconvex}. These simplified metrics can be easily calculated but fail to capture subtle effects. For example, the predicted values may have different variances across groups. Recently, researchers started to propose fairness metrics based on distributions/densities instead of simple point estimate~\cite{DBLP:conf/icml/AgarwalDW19}, and develop approximation methods \cite{DBLP:journals/corr/abs-2002-06200} for achieving fairness in regression. It is imperative to develop a general fair regression framework that enforces a variety of fairness notions and provides efficient implementation and theoretical analysis when dual optimization and approximation are applied.  Moreover, all previous fair regression research assumed the training data and testing data are drawn from the same distributions. This assumption is often violated in real world due to the sample selection bias between the training and testing data.

\begin{figure*}
\centering
\includegraphics[scale = 0.5]{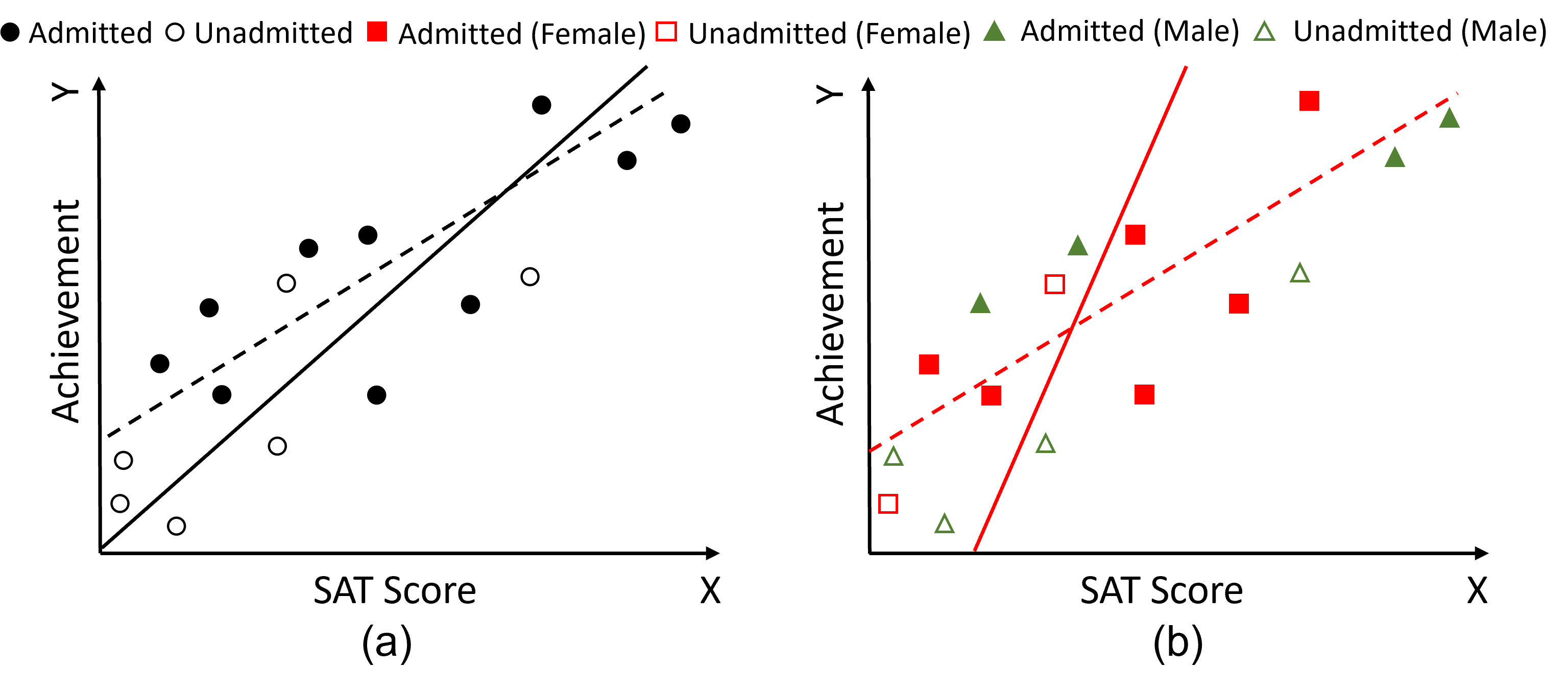}
\caption{Illustration for fair regression under sample selection bias. Fitted models with bias correction use feature values of unadmitted students.
\rule[.5ex]{1.5em}{.4pt} LR w/o  correction, \sampleline{dashed} LR with correction,
{\color{red}\rule[.5ex]{1.5em}{.4pt}}  fair LR w/o  correction,
{\color{red}\sampleline{dashed}} fair LR  with  correction}
\label{fig:framework}
\end{figure*}

Figure \ref{fig:framework} shows an illustrative example of studying the relationship between SAT scores ($X$) and potential college achievement ($Y$) of students. The regression model trained on only observed student samples who were already admitted to college (denoted as $\mathcal{D}_s$ and shown as solid data points in Figure \ref{fig:framework}(a)) would be biased as the fitted model did not consider applicants who could potentially go to college (denoted as $\mathcal{D}_u$ and shown as hollow points in Figure \ref{fig:framework}(a)). Note that for these applicants who did not go to college, SAT scores ($X$) are still available although their corresponding college achievements ($Y$) are missing.  Moreover, as shown in Figure \ref{fig:framework}(b), the fair regression model trained on only admitted college students $D_s$ in fact would be unfair and cannot be adopted for future applicants whose distribution is assumed to resemble the union of $D_s$ and $D_u$.  It is imperative to learn a fair regression model that can incorporate $X$ values of samples from  $\mathcal{D}_u$ to both improve model fitness and achieve fairness on population.

In this paper, we propose, $\textit{FairLR}^\star$, the fair regression framework under sample selection bias when dependent variable values of a set of samples from the training data are missing as a result of another hidden selection process. Our $\textit{FairLR}^\star$ adopts the classic Heckman model \cite{heckman1979sample} for bias correction and the Lagrange duality theory \cite{boyd2004convex}  to achieve regression fairness based on a variety of fairness notions.  Our fair regression framework minimizes the loss function subject to fairness inequality and equality constraints. We apply the Lagrange duality theory to transform the primal problem into a dual convex optimization problem.   For the two popular fairness notions, mean difference (MD) and mean squared  error difference (MSED), we derive two explicit formulas without optimizing iteratively. For Pearson correlation, we derive its conditions of satisfying the Slater condition, thus achieving strong duality. We conduct experiments on three real-world datasets and the experimental results demonstrate our approach's effectiveness in terms of both utility and fairness.

\section{Related Work}

{\noindent \bf Fair Regression.}
For linear regression $f(\cdot): X \rightarrow Y$ with discrete sensitive attribute $A$,  \cite{calders2013controlling} first introduced mean difference and AUC to measure the unfairness.   \cite{fitzsimons2019general} also used a similar concept termed as group fairness expectation to ensure fair prediction for different groups. For regression with discrete/continuous sensitive attribute,  \cite{mary2019fairness} used the R$\acute{e}$nyi maximum correlation coefficient of prediction and sensitive attribute to describe the fairness penalty.  Recently,  \cite{DBLP:conf/icml/AgarwalDW19} presented two fairness definitions, statistical parity and bounded group loss. The statistical parity uses the departure of the cumulative distribution function (CDF) of $f(X)$ conditional on $A=a$ from the CDF of $f(X)$. When the departure is close to zero, the prediction is statistically independent of the protected attribute. The bounded group loss requires that the prediction error of any protected group stay below some pre-determined thresholds.

To address the challenge of estimating information-theoretic divergences between conditional probability density functions,  \cite{DBLP:journals/corr/abs-2002-06200} introduced fast approximations of the independence, separation and sufficiency group fairness criteria for regression models from their (conditional) mutual information definitions.
 \cite{chzhen2020fair} focused on demographic parity that requires the distribution of the predicted output  independent of the sensitive attribute.  They established a connection between fair regression and optimal transport theory and derived a closed form expression for the optimal fair predictor, i.e., the distribution of this optimum is the Wasserstein barycenter of the distributions induced by the standard regression function on the sensitive groups.

{\noindent \bf Fair Classification under Sample Selection Bias.} The sample selection bias causes the training data to be selected non-uniformly from the population to be modeled. Generally there are four types of sample selection bias, missing completely at random, missing at random, missing at random-class, and missing not at random when there is no independence assumption between features, target, and selection.
Extensive research has been conducted on classification under sample selection bias (refer to a survey \cite{moreno2012unifying}).
Some recent research focused on robust classification under sample selection bias and covariate shift.
For example,  \cite{wen2014robust} considered covariate shift between the training and testing data and proposed a minimax robust framework that applies Gaussian kernel functions to reweigh the training examples.  \cite{du2021robust} adopted the reweighing estimation idea for sample selection bias correction and used the minimax robust estimation to achieve robustness on prediction accuracy.
To tackle the unfairness issue under distribution shift,  \cite{taskesen2020distributionally} developed a distributionally robust logistic regression model with an unfairness penalty. They assumed the unknown true test distribution is contained in a Wasserstein ball centered at the  empirical distribution on the observed training data.
\cite{DBLP:conf/aaai/RezaeiFMZ20} proposed the use of ambiguity set to derive the fair classifier based on the principles of distributional robustness. There are also other related works including fair transfer learning \cite{schumann2019transfer}, fair federated learning \cite{du2021fairness} and fair classification with bias in label collection \cite{kallus2018residual,jiang2020identifying}.

\section{Fair Regression under Sample Selection Bias}

\subsection{Problem Formulation}
We  first define the notations used in this paper. Let $(\bm{X}, A, Y)$ denote the training data $\mathcal{D}$, where $\bm{X}$ is the feature space, $A$ is the protected attribute, and $Y$ is the continuous target attribute. $\mathcal{D}$ contains $n$ data samples, among which $m$ samples $(\bm{x}_i, a_i, y_i)$ are fully observed and the remaining $n-m$ data points have $y_i$ missing. We denote the fully observed part as $\mathcal{D}_s$ and the other part as $\mathcal{D}_u$. The whole training data $\mathcal{D}$ is selected uniformly from the population to be modeled. However,  $\mathcal{D}_s$ is  non-uniformly selected  and the bias of $\mathcal{D}_s$ could depend on both feature vector $\bm{x},a$ and target variable $y$. A regression function $f(\cdot): X \rightarrow Y$ tries to learn optimal parameter $\bm{w}$. We denote $\hat{y}= f(\bm{x}; \bm{w})$.

\noindent \textbf{Problem Statement.} Given the training dataset $\mathcal{D} = \mathcal{D}_s \cup \mathcal{D}_u$,
derive regression function $f(\cdot): X \rightarrow Y$ that achieves fairness on population with respect to protected attribute $A$ based on some fairness criterion.

We emphasize the sample selection bias considered in our paper is different from the traditional covariate shift scenario. Although the covariate shift tackles the shift between the training distribution $P_{tr}(X)$ and test distribution $P_{te}(X)$, it usually assumes both a labeled training dataset and an unlabeled testing dataset are available in the training phase. In our work, we do not require the unlabeled testing data in the training phase. Instead, we assume the available training dataset contains a mixture of labeled and unlabeled data points but the labeling process is biased.
In our setting, we are able to use the Heckman model to correct the  bias with theoretical guarantee as we can compute the conditional unbiased expectation analytically. However, the previous commonly used approaches for covariate shift can only achieve robust estimation within a range but cannot provide theoretical guarantee.

\subsection{Heckman Model Revisited}

Heckman model \cite{heckman1979sample} addresses the issue of sample selection bias when the dependent variable in the regression has values that are missing not at random.  In the two-step estimation procedure of Heckman model, the first step uses probit regression to model the sample selection process and derives a new variable called the Inverse Mills Ratio (IMR). The second step adds the IMR to the regression analysis as an independent variable and uses ordinary least squares to estimate the regression coefficients. This two-step estimator can perform well when there is no multicollinearity between the IMR and the explanatory variables. We present below the Heckman model formally.

The selection equation of the $i$th sample is $z_i = \bm{x}_{1i}\bm{\gamma} + u_i$
where $\bm{x}_{1i}$ includes the set of features related to sample selection, $\bm{\gamma}$ is the set of regression coefficients, and $u_i$ is the error term. The selection index $s$ is defined as:
\begin{equation}
s_i =\left\{
\begin{aligned}
&1 \hspace{0.5cm} z_i > 0 \\
&0  \hspace{0.5cm} z_i \le 0 \\
\end{aligned}
\right.
\end{equation}
where $s_i = 1$ indicates that the $i$th sample is fully observed and $s_i = 0$ indicates its target value $y_i$ is missing.
The prediction model is based on linear regression and for the $i$th sample we have $y_i = \hat{y}_i+ \epsilon_i = {\bm{x}}_{2i}\bm{\beta} + \epsilon_i$
where $\hat{y}_i$ is the predicted value, ${\bm{x}}_{2i}$ includes the set of features used for prediction, $\bm{\beta}$ is the set of regression coefficients, and $\epsilon_i$ is the error term. Following the default assumptions in the Heckman model, ${\bm{x}}_{2i}$ is a subset of $\bm{x}_{1i}$, indicating that all attributes predicting the outcome of interest can also predict selection equation, and  $u_i \sim N(0, 1)$, and $\epsilon_i \sim N(0, \sigma_{\epsilon}^2)$. The correlation coefficient of $u_i$ and $\epsilon_i$ is denoted by $\rho$.
The prediction outcome based on $\mathcal{D}_s$ alone is biased and we can correct it by computing the conditional means of the prediction outcome as:
\begin{equation}\label{eq: unbiased_expectation}
\scalebox{0.85}{
$\begin{aligned}
  \mathbb{E}(y_i|s_i = 1) = \mathbb{E}(y_i|z_i > 0) &= \mathbb{E}( {\bm{x}}_{2i}\bm{\beta} + \epsilon_i|\bm{x}_{1i}\bm{\gamma} + u_i > 0) \\
 &= {\bm{x}}_{2i}\bm{\beta} + \mathbb{E}(\epsilon_i|\bm{x}_{1i}\bm{\gamma} + u_i > 0) \\
 &= {\bm{x}}_{2i}\bm{\beta} + \mathbb{E}(\epsilon_i|u_i > -\bm{x}_{1i}\bm{\gamma})
\end{aligned}$}
\end{equation}
Because $u_i$ and $\epsilon_i$ are correlated, then we have (see the supplemental material for derivations)
\begin{equation}
    \mathbb{E}(\epsilon_i|u_i > -\bm{x}_{1i}\bm{\gamma}) = \alpha_i \rho \sigma_{\epsilon}
\end{equation}
where $\alpha_i = \dfrac{\phi(-\bm{x}_{1i}\bm{\gamma})}{1 - \Phi(-\bm{x}_{1i}\bm{\gamma})}$ is usually termed as IMR. Here  $\phi(\cdot)$ denotes the standard normal density function and $\Phi(\cdot)$ denotes the standard cumulative distribution function.

To compute the value of $\alpha_i$, the first step is to estimate the coefficients $\bm{\gamma}$. We use the maximum likelihood estimate (MLE) to estimate $\bm{\gamma}$ by treating the selection equation as a probit classification model and we have
\begin{equation}
    P(s_i = 1) = \Phi (\bm{x}_{1i}\bm{\gamma}), P(s_i = 0) = 1- \Phi (\bm{x}_{1i}\bm{\gamma})
\end{equation}
Then the likelihood of $\mathcal{D}$ is expressed as:
\begin{equation}\label{eq: mle}
    LH(\bm{\gamma}; s_i, \bm{x}_{1i}) = \prod_{i=1}^{n}  \Phi (\bm{x}_{1i}\bm{\gamma})^{s_i}(1-\Phi (\bm{x}_{1i}\bm{\gamma}))^{1-s_i}
\end{equation}
The maximization of Eq. \ref{eq: mle} will obtain the estimates of $\bm{\gamma}$, and thus we can compute $\alpha_i$ for each selected sample $(x_i, a_i, y_i)$ in $D_s$.
With available $\alpha_i$, we can rewrite Eq. \ref{eq: unbiased_expectation} as:
\begin{equation}\label{eq: organized_expectation}
     \mathbb{E}(y_i|s_i = 1) = {\bm{x}}_{2i}\bm{\beta} +\alpha_i \rho \sigma_{\epsilon}
\end{equation}
Then we can estimate the coefficients $\bm{\beta}$ from Eq. \ref{eq: organized_expectation}, e.g., via the ordinary least squares (OLS) by minimizing  $\min_{\bf{\beta}}L(\bm{\beta})
    = \sum_{i = 1}^{m}(\bm{x}_{2i}\bm{\beta} +\alpha_i \rho \sigma_{\epsilon}  - y_i)^2$.

\subsection{Fair Regression via Heckman Correction}

We first present the general fair regression framework that aims to minimize the risk and learns the parameters $\bm{w} \in \mathcal{W}$  subject to the fairness constraints:
\begin{equation}\label{eq:fml}
\begin{aligned}
&\min_{\bm{w}\in \mathcal{W}} \mathbb{E}[l(\hat{y}, y)] = \mathbb{E}[l(f(\bm{x};\bm{w}), y)]\\
&\text{subject to} \quad g_i(\hat{y}, y, a) \leq 0, i=1,\cdots,p\\
 & \quad\quad\quad\quad\quad h_j(\hat{y}, y, a) = 0, j=1,\cdots,q
\end{aligned}
\end{equation}
where $f$ is the learning model, $l$ is the loss function, $g_i$  are fairness inequality  constraints, and $h_j$ are fairness equality constraints.

We then formulate the fair regression under sample selection bias and rewrite Eq. \ref{eq:fml} to minimize the empirical loss subject to the fairness constraint:
\begin{equation}\label{eq:fheckman}
\begin{aligned}
&p^*=\min_{\tilde{\bf{\beta}}}L(\tilde{\bm{\beta}}) = \sum_{i=1}^m{l[(f_h(\tilde{\bm{x}}_{2i}; \tilde{\bm{\beta}}), y)]}
\\
&\text{subject to}\quad g_i(\tilde{\bm{\beta}}) \leq 0, i=1,\cdots,p\\
 & \quad\quad\quad\quad\quad h_j(\tilde{\bm{\beta}}) = 0, j=1,\cdots,q
\end{aligned}
\end{equation}
where $\tilde{\bm{\beta}} = [\bm{\beta}, \bm{\beta}_{\alpha}]$, $\beta_{\alpha}$ = $\rho \sigma_{\epsilon}$, $\tilde{\bm{x}}_{2i} = [\bm{x}_{2i}, \alpha_i]$, and
$f_h(\tilde{\bm{x}}_{2i}; \tilde{\bm{\beta}}) = \tilde{\bm{x}}_{2i}\tilde{\bm{\beta}}$ is the Heckman prediction function.
Note that from Eq. \ref{eq: organized_expectation} the bias is corrected by $\alpha_i$ which carries the information of $\mathcal{D}_u$. The effect of each $\alpha_i$ on the sample $(x_i, a_i, y_i)$ is quantified by $\rho \sigma_{\epsilon}$. We can then treat $\rho \sigma_{\epsilon}$  as one additional dimension $\beta_{\alpha}$ of the coefficient vector  $\tilde{\bm{\beta}}$. In addition, the fairness constraint computed based on the corrected $\tilde{\bm{\beta}}$ and $ \tilde{\bm{x}}_{2i}$ is unbiased.

\subsubsection{Fairness Notions}
Previous works on fair regression developed notions are based on the independence of model prediction (or prediction error) and sensitive attribute. Different from the classification setting,  target variable $Y$ is continuous and sensitive attribute $A$ can be either categorical or continuous. Table \ref{tab:crime_defintion1} summarizes fairness notions including their formula, reference, and applicability in terms of sensitive attribute type. We leave their formal definitions in the supplemental material.  The mean difference (MD),  the mean squared  error difference (MSED), the statistical parity (SP), and the bounded group loss (BGL) handle categorical sensitive attribute whereas Pearson correlation ($\rho_{\hat{y}a}$) and our introduced
partial correlation ($\rho_{\hat{y}a.y}$) handles numerical sensitive attribute. The partial correlation $\rho_{\hat{y}a.y}$ includes both $y$ and $a$ in the condition which is similar to the equalized opportunity \cite{hardt2016equality} in fair classification.
MD, Pearson and SP focus on independence of model prediction and sensitive attribute whereas MSED, Partial and BGL consider prediction error. Moreover, SP (BGL) measures the dependence of prediction (prediction error) and sensitive attribute on distributions/densities, in contrary to point estimate of other notions.

\begin{table}[htb]
	\caption{Fairness Notions for Regression}
	\centering
	\renewcommand{\arraystretch}{2}
	\scalebox{0.8}{
	\begin{tabular}{cccccc}
		\hline
		\textbf{Definition} & Reference & Equation & Categorical & Numeric \\
		\hline
		MD &  \cite{calders2013controlling, johnson2016impartial,  berk2017convex, zhao2020unfairness} &  $\text{MD}(\hat{y}, a) = \mathbb{E}(\hat{y}|a = 0) - \mathbb{E}(\hat{y}|a = 1)$
		& $\checkmark$ & $\times$   \\
		\hline
	    MSED & \cite{calders2013controlling, johnson2016impartial} & $\text{MSED}(\hat{y}, a) =\mathbb{E}[(y - \hat{y})^2|a = 0] - \mathbb{E}[(y - \hat{y})^2|a = 1] $ & $\checkmark$ & $\times$  \\
		\hline
		Pearson & \cite{komiyama2018nonconvex} & $\rho_{\hat{y}a} = \frac{\mathbb{E}[(\hat{y} - \mu_{\hat{y}})(a - \mu_a)]}{\sigma_{\hat{y}} \sigma_{s}}$ &  $\times$ & $\checkmark$\\
		\hline
		Partial & ours  &  $\rho_{\hat{y}a.y} = \frac{\rho_{\hat{y}a}-  \rho_{\hat{y}y}  \rho_{ay}}{\sqrt{1-\rho^2_{\hat{y}{y}}}\sqrt{1-\rho^2_{ay}}}$  &  $\times$ & $\checkmark$ \\
		\hline
		SP & \cite{DBLP:conf/icml/AgarwalDW19, narasimhan2020pairwise, le2020computing, DBLP:conf/nips/ChzhenDHOP20a} & $\text{SP} = \mathbb{P}[f(X) \ge z|A=a] - \mathbb{P}[f(X) \ge z] $  &  $\checkmark$ & $\times$\\
		\hline
		BGL & \cite{DBLP:conf/icml/AgarwalDW19} & $ \text{BGL} = \mathbb{E}[l (f(X),Y)|A=a]$ &  $\checkmark$ & $\times$\\
		\hline
	\end{tabular}}\label{tab:crime_defintion1}
\end{table}

In general, we can enforce strict fairness via equality constraints and relaxed fairness via inequality constraints in Eq. \ref{eq:fheckman}. For example,  we use $h_j(\tilde{\bm{\beta}}) =0$ for $\text{MD} = 0$, and  use $g_i(\tilde{\bm{\beta}}) - \tau \le 0$ for $\text{MD} \le \tau$ where $\tau$ is a user-specified  threshold. One challenge is for SP as the number of constraints is uncountable. We can apply the algorithm developed in \cite{DBLP:conf/icml/AgarwalDW19} that discretizes the real-valued prediction space and reduces the optimization problem to cost-sensitive classification. The cost-sensitive classification is then solved by the reduction approach \cite{agarwal2018reductions}. We note that our framework can be used to enforce multiple fairness notions at the same time and some notions may be mutually contradictory \cite{kleinberg2016inherent}, which can cause vacuous solutions.

\subsubsection{Dual Formulation}

To solve the primal optimal problem (Eq. \ref{eq:fheckman}) with a variety of fairness notions, we apply the Lagrange duality theory \cite{boyd2004convex} to relax the primal problem by its constraints. The Lagrangian function is
\begin{equation}
L_c(\tilde{\mathbf{\beta}}, \mathbf{\lambda}, \mathbf{\upsilon}) = L(\tilde{\mathbf{\beta}}) + \mathbf{\lambda}^T g(\tilde{\mathbf{\beta}})  + \mathbf{\upsilon}^T h(\tilde{\mathbf{\beta}})
\end{equation}
where $\mathbf{\lambda} \in \mathbb{R}_{+}^p$ and $\mathbf{\upsilon} \in \mathbb{R}^q$  are the Lagrange multiplier vectors (or dual variables) associated with inequality constraints and equality constraints. The dual function hence is defined as $Q(\mathbf{\lambda}, \mathbf{\upsilon}) = \inf_{\tilde{\mathbf{\beta}}} {L_c(\tilde{\mathbf{\beta}}, \mathbf{\lambda}, \mathbf{\upsilon})}$.
Note that the dual function $Q(\mathbf{\lambda}, \mathbf{\upsilon})$ is a pointwise affine function of $(\mathbf{\lambda},\mathbf{\upsilon})$, it is concave even when the problem (Eq. \ref{eq:fheckman}) is non-convex.
For each pair $(\mathbf{\lambda},\mathbf{\upsilon})$, the dual function gives us a lower bound of the optimal value $p^*$, i.e., $Q(\mathbf{\lambda}, \mathbf{\upsilon}) \le p^*$.
The best lower bound leads to the Lagrange dual problem:
\begin{equation}\label{eq:dual-full}
\begin{aligned}
d^*=\max_{\mathbf{\lambda} \succeq 0, \mathbf{\upsilon}}  Q(\mathbf{\lambda}, \mathbf{\upsilon}) = \max_{\mathbf{\lambda} \succeq 0, \mathbf{\upsilon}}  \min_{\tilde{\mathbf{\beta}}} {L_c(\tilde{\mathbf{\beta}}, \mathbf{\lambda}, \mathbf{\upsilon})}
\end{aligned}
\end{equation}
The Lagrange dual problem is  a convex optimization problem because the objective to be maximized is concave and the constraint is convex. We can solve the dual optimization problem by alternating gradient descent steps over the primal variables $\tilde{\mathbf{\beta}}$ and dual variables $(\mathbf{\lambda}, \mathbf{\upsilon})$, respectively. In particular, by iteratively executing the following two steps: 1) find  $\tilde{\mathbf{\beta}}^{*} \leftarrow argmin_{\tilde{\mathbf{\beta}}}  {L_c(\tilde{\mathbf{\beta}}, \mathbf{\lambda}, \mathbf{\upsilon})}$; 2) compute $\mathbf{\lambda} \leftarrow \mathbf{\lambda} + \eta \frac{dL_c}{d{\mathbf{\lambda}}} (\tilde{\mathbf{\beta}}^{*}, \mathbf{\lambda}, \mathbf{\upsilon})$, $\mathbf{\upsilon} \leftarrow \mathbf{\upsilon} + \eta \frac{dL_c}{d{\mathbf{\upsilon}}} (\tilde{\mathbf{\beta}}^{*}, \mathbf{\lambda}, \mathbf{\upsilon})$, the solution will converge.

Next, we show instantiations of two widely used fairness notions, MD and MSED, by deriving their explicit formulas without iterative optimization. We leave the detailed proofs in the supplementary material.

\begin{result}\label{result:MD}
For fair regression with the mean squared loss and $\text{MD}(\hat{y},a) =0$, we have the closed solution
\begin{equation}\label{eq:MD_Solution}
    \tilde{\bm{\beta}} = (\tilde{\bm{X}}_2^T\tilde{\bm{X}}_2)^{-1} (\tilde{\bm{X}}_2^T\bm{y}-\frac{\bm{d}^T(\tilde{\bm{X}}_2^T\tilde{\bm{X}}_2)^{-1} \tilde{\bm{X}}_2^T\bm{y}}{\bm{d}^T(\tilde{\bm{X}}_2^T\tilde{\bm{X}}_2)^{-1} \bm{d}} \bm{d})
\end{equation}
\end{result}

\begin{proof}[Proof Sketch]
The dual optimization form is
$L(\tilde{\bm{\beta}}) = \min \sum_{i = 1}^{m}(\tilde{\bm{\beta}}\tilde{\bm{x}}_{2i} - y_i)^2 + 2\lambda \bm{d}^T\tilde{\bm{\beta}}$
where $\bm{d} = \dfrac{1}{m_0}{\sum_{i\in \mathcal{D}_0}\tilde{\bm{x}}_{2i}} - \dfrac{1}{m_1}{\sum_{i \in \mathcal{D}_1}\tilde{\bm{x}}_{2i}}$,  $m_0$ ($m_1$) is the number of data in $\mathcal{D}_s$ with $a = 0$ ($1$).
By setting the derivative of $L(\tilde{\bm{\beta}})$ with respect to $\tilde{\bm{\beta}}$ be zero, we get $\tilde{\bm{\beta}} = (\tilde{\bm{X}}_2^T\tilde{\bm{X}}_2)^{-1} (\tilde{\bm{X}}_2^T\bm{y}-\lambda \bm{d})$, where $\tilde{\bm{X}}_2$ is the matrix form of $\tilde{\bm{x}}_{2i}, i \in [m]$ and $\bm{y}$ is the vector form of $y_i, i \in [m]$. Using the fairness constraint, we  get the closed solution of $\lambda$ and then $\tilde{\bm{\beta}}$ as Eq. \ref{eq:MD_Solution}.
\end{proof}

\begin{result}\label{result:MSED}
For fair regression with the mean squared loss and $\text{MSED}(\hat{y}, a) = 0$, we have the closed solution
\begin{equation}
\begin{aligned}
\tilde{\bm{\beta}} = &(\tilde{\bm{X}}_{2}^T\tilde{\bm{X}}_2 + \dfrac{\lambda}{m_0}(\tilde{\bm{X}}_{2}^0)^T\tilde{\bm{X}}_2^0 - \dfrac{\lambda}{m_1}(\tilde{\bm{X}}_{2}^1)^T\tilde{\bm{X}}_2^1)^{-1} \\&(\tilde{\bm{X}}_{2}^T\bm{y} + \dfrac{\lambda}{m_0}(\tilde{\bm{X}}_{2}^0)^T\bm{y}_0 - \dfrac{\lambda}{m_1}(\tilde{\bm{X}}_{2}^1)^T\bm{y}_1)
\end{aligned}
\end{equation}
\end{result}

\begin{proof}[Proof Sketch] Similar to Result \ref{result:MD}, we  write its Lagrange dual form of the MSED fairness constraint. Then we set the derivative of $\tilde{\bm{\beta}}$ to be zero, and compute the solution of $\lambda$ and $\tilde{\bm{\beta}}$. Note that $\tilde{\bm{X}}_2^0$ is the matrix form of $\tilde{\bm{x}}_{2i}, i \in [m_0]$, $\tilde{\bm{X}}_2^1$ is the matrix form of $\tilde{\bm{x}}_{2i}, i \in [m_1]$, $\bm{y}_0$ is the vector form of $y_i, i \in [m_0]$, and $\bm{y}_1$ is the vector form of $y_i, i \in [m_1]$.
\end{proof}

\subsubsection{Duality Gap Analysis}
The optimal value $d^*$ of the Lagrange dual problem, by definition, is the best lower bound on $p^*$ that can be obtained from the Lagrange dual function. The difference $p^*-d^*$, which is always nonnegative, is the optimal duality gap of the original problem.
One theoretical question is whether and under what conditions we can achieve zero duality gap (i.e., the optimal values of the primal and dual problems are equal) in our fair regression framework.

\begin{result}
\label{result:slate}
For fair regression with the convex loss function and
the fairness inequality constraints (i.e., less than a user-specified threshold $\tau$), the strong duality holds for Pearson correlation if the linear relationship exists between $x$ and $a$.
\end{result}

\begin{proof}[Proof Sketch]
Our proof is based on strong duality via Slater condition \cite{boyd2004convex}.
We leave the detailed proof in the supplemental material and provide the proof sketch below.
The correlation between $\tilde{\bm{x}}_2$ and $a$ can be removed via the following:
\begin{equation}\label{Eq:correlation_remove}
    \hat{\bm{B}} = (\bm{A}^T\bm{A})^{-1}\tilde{\bm{X}}_{2}, \bm{U}=\tilde{\bm{X}}_{2} - \hat{\bm{B}}\bm{A}
\end{equation}
where $\bm{A} = (a_1, a_2, \cdots, a_n)$ and we define $\bm{u}_i$ as the $i$-th data point of $\bm{U}$. Then we can compute the Pearson correlation between $\hat{y}$ and $a$ as the following:
\begin{equation}
\begin{aligned}
\rho(\hat{y}, a)=\dfrac{\tilde{\bm{\beta}}_a\sqrt{Var(a)} }{\sqrt{\tilde{\bm{\beta}}_a^2Var(a) + \tilde{\bm{\beta}}_u^T \bm{V}_u\tilde{\bm{\beta}}_u}}
\end{aligned}
\end{equation}
where $Var(a)$ ($\bm{V}_u$) is the covariance of $a$ ($\bm{u}$) and $\tilde{\bm{\beta}}_a$ ($\tilde{\bm{\beta}}_u$) corresponds to the coefficient of $\tilde{\bm{\beta}}$  with respect to $a$ ($\bm{u}$). Then the fairness constraint of Pearson correlation  $\rho^2(\hat{y}, a) \leq \epsilon$ is equivalent to:
\begin{equation}
    (1 - \epsilon)\tilde{\bm{\beta}}_a^2Var(a) - \epsilon \tilde{\bm{\beta}}_u^T \bm{V}_u\tilde{\bm{\beta}}_u \leq 0
\end{equation}
It can be verified that $\{(\tilde{\bm{\beta}}_a, \tilde{\bm{\beta}}_u): (1 - \epsilon)\tilde{\bm{\beta}}_a^2Var(a)  - \epsilon \tilde{\bm{\beta}}_u^T \bm{V}_u\tilde{\bm{\beta}}_u < 0\} \neq  \emptyset$. Therefore, it satisfies the requirement of Slater Condition and thus the strong duality holds.
\end{proof}

{\bf \noindent Remarks.} Note that we do not need to conduct the duality gap analysis for the mean difference (MD) and the mean squared error difference (MSED) as Results \ref{result:MD} and
\ref{result:MSED} have already given the explicit formulas for the primal optimization. For Partial, SP, BGL and other potential fairness notions, we leave their analysis in our future work. Moreover, when there is no sample selection bias, our Results 1-3 naturally hold by removing the tilde from those tilde symbols (e.g., $\tilde{\bm{\beta}}$).

\section{Experiments}

\subsection{Experiment Setting}
\textbf{Datasets.} We conduct our experiment on three real-world datasets that are widely used to evaluate fair machine learning models. For each dataset, we choose 70\% of data as training data $\mathcal{D}$ and leave the rest as testing data. To create the sample selection bias, we follow the procedure in \cite{laforgue2019statistical} by splitting $\mathcal{D}$ into $\mathcal{D}_s$ (samples with fully observed features $X$ and target $Y$) and $\mathcal{D}_u$ (samples with missing $Y$) according to some specific features. We show the characteristics of three datasets including protected attribute $A$, target $Y$,  sizes of $\mathcal{D}_s$, $\mathcal{D}_u$, and testing data in Table \ref{tab:dataset} and show the attribute lists used in selection/prediction in the supplementary material.

\begin{table}[h]
	\caption{Characteristics of datasets}
	\centering
	\scalebox{0.8}{
	\begin{tabular}{|c|c|c|c|c|c|}
		\hline
		\textbf{Dataset} & Protected $A$ & Target $Y$ & $|\mathcal{D}_s|$  & $|\mathcal{D}_u|$  & Testing \\
		\hline
		CRIME & AAPR  & Crime Rate & 976 & 419 & 599\\
	   \hline
	   LAW &  Black/Non-black & GPA &  1323 & 567 & 810 \\
		\hline
		COMPAS &  Black/Non-black & Risk Score &  2153 & 924 & 1320 \\
		\hline
	\end{tabular}}\label{tab:dataset}
\end{table}

\begin{figure*}[htbp]
  \centering
  \includegraphics[scale = 0.7]{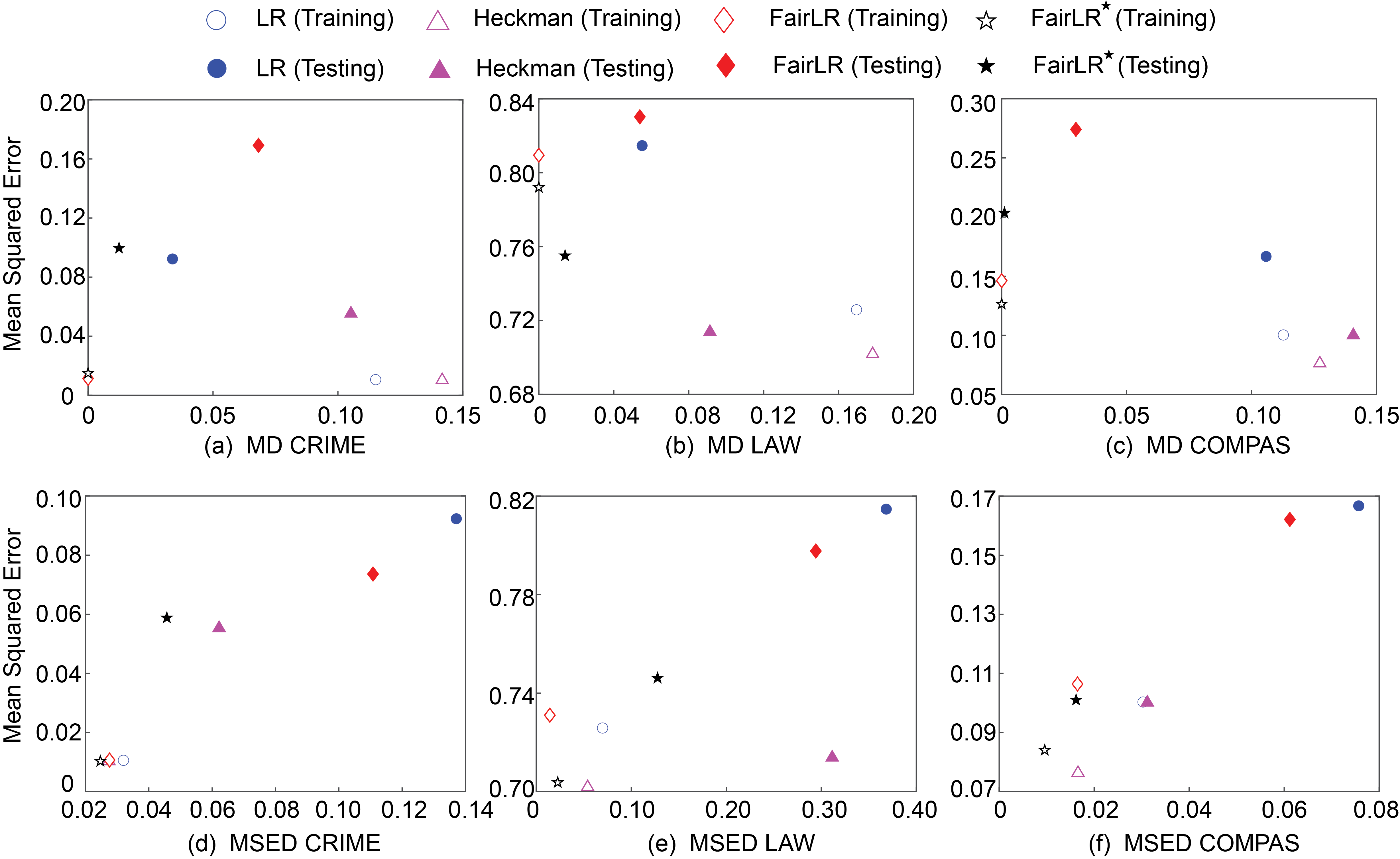}
  \caption{Performance evaluation of binary protected attribute on CRIME, LAW, and COMPAS. The closer to the origin, the better the accuracy-fairness trade-off.}\label{fig:binary}
\end{figure*}
CRIME dataset~\cite{Doe:2009:Online}  was collected from the 1990 US Census and contains socio-economic data of 1994 communities.
The task is to predict the crime rate of a given community based on its socio-economic information. We choose the African American Population Ratio (AAPR) as the sensitive attribute and label a community as protected if its AAPR is greater than 50\% and non-protected otherwise. In total, we have 219 protected communities and 1775 non-protected communities. In our experiments, we remove attributes with missing values and standardize all attributes to have zero mean value and unit variance.  We include samples to $\mathcal{D}_s$ if the ratio of people under the poverty level in a community is less than 0.05, and samples to $\mathcal{D}_u$ otherwise.

LAW dataset \cite{wightman1998lsac} was collected from the Law School Admissions Council’s National Longitudinal Bar Passage Study and consists of personal records of law students who went on to take the bar exam, including LSAT score, age, race and so forth. The task is to predict the GPA of a student based on other attributes. We choose race as the sensitive attribute and treat  black as protected. The dataset contains a total of 20649 records and we randomly select 2700 records, including 700 protected samples and 2000 non-protected samples. We include samples to $\mathcal{D}_s$ if the year of birth is after 1950, and samples to $\mathcal{D}_u$ otherwise.

COMPAS dataset \cite{compass} consists of a collection of data from criminal defenders from Florida in 2013-2014. Each data sample is associated with personal information, including race, gender, age, prior criminal history, and so forth. The task is to predict the risk level of a defender based on other attributes. We choose race as the sensitive attribute and treat black defenders as protected. After removing the duplicated data samples, we have a total of 4397 data samples, including 2694 protected samples and 1703 non-protected samples. We include samples to $\mathcal{D}_s$ if the year of decile score  is less than 10, and to $\mathcal{D}_u$ otherwise.

\textbf{Baseline Models and Metrics.} We choose linear regression with the standard loss function, mean squared loss, in our proposed framework $\textit{FairLR}^\star$. We adopt each of four fairness metrics,  MD, MSED, Pearson coefficient and Partial coefficient, with equality constraint forms. We consider the following baseline models: (a) Linear regression (\textit{LR}) without fairness constraint; (b) Linear regression with Heckman correction (\textit{Heckman}) from \cite{heckman1979sample} (c) Linear regression with each fairness constraint (\textit{FairLR}), including MD \cite{calders2013controlling}, MSED \cite{johnson2016impartial}, Pearson coefficient \cite{komiyama2018nonconvex}, and Partial coefficient.  We evaluate the performance of the proposed framework based on prediction accuracy and fairness.  We use the mean squared error (MSE) to measure prediction accuracy. For fairness, we use MD and MSED in the  binary sensitive attribute setting and  Pearson coefficient and Partial coefficient in the numerical sensitive attribute setting. As the goal of fair regression is to achieve good accuracy and fairness on population, we use  MSE and fairness calculated from testing data to compare different models. For a comprehensive comparison, we also report those values calculated from $\mathcal{D}_s$.

\subsection{Evaluation on Binary Protected Attribute}\label{sec: binary}

We report in Figure \ref{fig:binary} our main comparison results on three datasets. Y-axis is MSE to reflect prediction accuracy and X-axis is based on the fairness metric chosen in fair regression models (\textit{FairLR} and $\textit{FairLR}^\star$). In particular, the three plots in the first row of Figure \ref{fig:binary} report MD whereas those on the second row report MSED. In each plot, we have eight markers with different shape and color, each of which reflects the MSE and fairness metric for one of the four compared models on either $\mathcal{D}_s$ or testing data. Throughout this section, we  use $\circ$, $\triangle$, $\diamond$, and $\star$ to denote  \textit{LR}, \textit{Heckman}, \textit{FairLR} and $\textit{FairLR}^\star$, and use hollow (solid) marker to represent results on $\mathcal{D}_s$ (testing data). In general, markers in bottom-left region (close to origin) indicate good performance of  corresponding methods as we want to achieve both low MSE for prediction and low MD/MSED for fairness.

\begin{figure}[t]
  \centering
  \includegraphics[scale = 0.6]{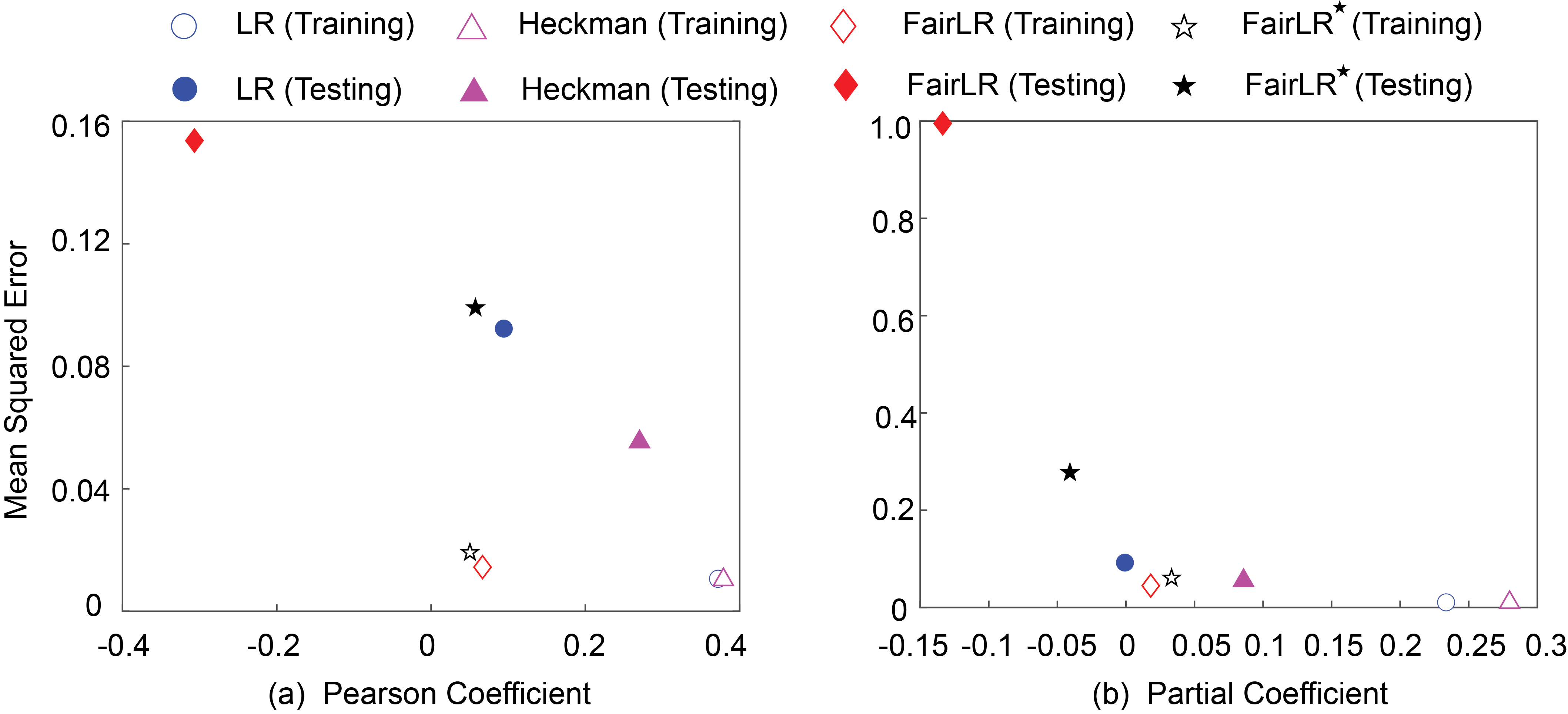}
  \caption{Performance evaluation of numerical protected attribute on CRIME.  The closer to the origin, the better the accuracy-fairness trade-off.} \label{fig:numerical}
\end{figure}

\begin{figure}[t]
  \centering
  \includegraphics[scale = 0.6]{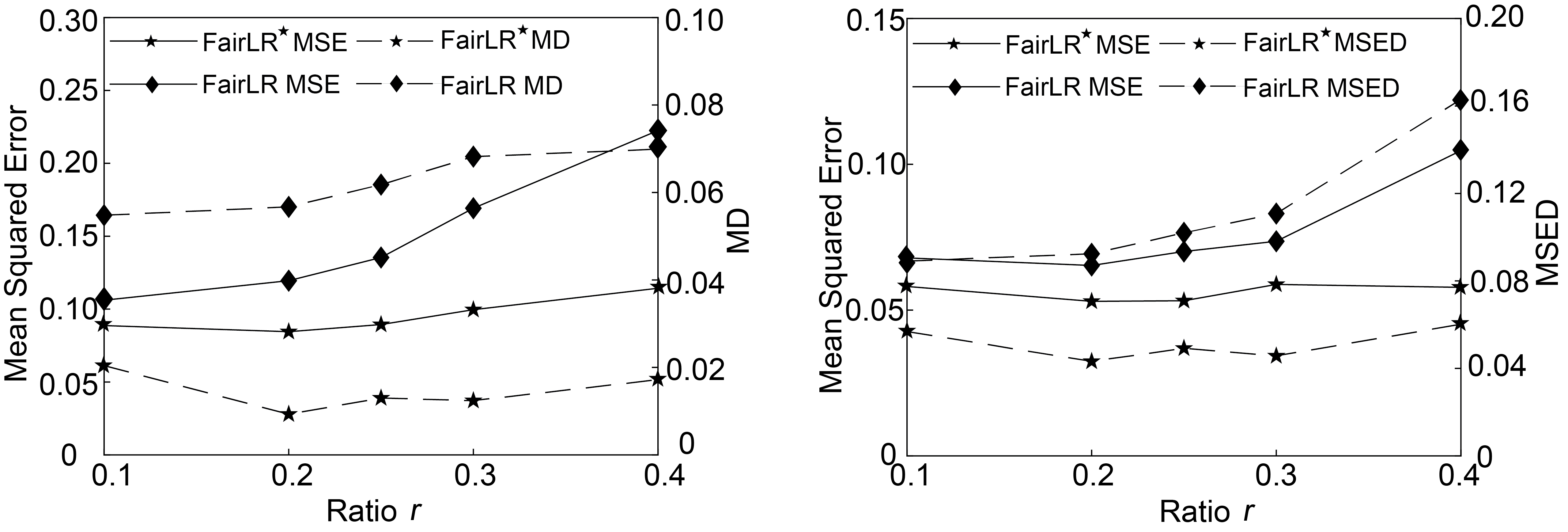}
  \caption{Effects of Ratio $r =|\mathcal{D}_u|/|\mathcal{D}|$}\label{fig:curve}
\end{figure}

We focus on the main results of comparing four methods on testing data, reflected by four solid markers in each plot~\footnote{Due to space limit, we skip the comparison results of four methods on $\mathcal{D}_s$ reflected by four hollow markers in each plot.}. We clearly see that markers of \textit{Heckman} always locate below that of \textit{LR}, indicating \textit{Heckman} successfully corrects the sample selection bias and reduces prediction error on testing data. Taking Figure \ref{fig:binary} (a) as an example, the MSE of \textit{Heckman} is 0.0553  whereas the MSE of \textit{LR} is 0.0923.  Similarly, as \textit{FairLR} does not do bias correction,  the solid marker of \textit{FairLR} is also higher than that of $\textit{FairLR}^\star$ for all three datasets. This demonstrates the effectiveness of Heckman model for correcting sample selection bias. Moreover, the solid marker of \textit{FairLR} is always on the right side of $\textit{FairLR}^\star$, reflecting that \textit{FairLR} simply trained on $\mathcal{D}_s$ without bias correction fails to achieve fairness on testing data. For example, in Figure \ref{fig:binary} (f),  the MSED of \textit{FairLR} is 0.0612  whereas that of $\textit{FairLR}^\star$ is 0.0162.  To conclude, our proposed $\textit{FairLR}^\star$ achieves the best trade-off between fairness and regression accuracy on the testing data.

It is also interesting to compare each model's performance between the training data and testing data. For our $\textit{FairLR}^\star$, we can see its hollow marker and solid marker are close to each other horizontally, indicating that the fairness achieved on $\mathcal{D}_s$ can also guarantee the testing fairness. However, the hollow marker and solid marker of \textit{FairLR} are separate, indicating the sample selection bias can incur unfairness in the testing data although \textit{FairLR} achieved  training fairness.

\subsection{Evaluation on Numerical Protected Attribute}

We conduct experiments on CRIME by using the original numerical attribute AAPR as sensitive attribute. We use Pearson coefficient to measure the independence between $\hat{Y}$ and $A$, and use Partial coefficient to measure the conditional independence of $\hat{Y}$ and $A$ given the true target value $Y$.  Figure \ref{fig:numerical} shows the comparison results of four models. We have similar observations as the binary protected attribute setting.
First, for results on $\mathcal{D}_s$ reflected by hollow markers, there is no surprise that all of the hollow markers are in bottom region with low MSE, and both \textit{FairLR} and $\textit{FairLR}^\star$ can achieve training fairness in terms of both Pearson coefficient and Partial coefficient. Second, for results on testing data reflected by solid markers, \textit{Heckman} ($\textit{FairLR}^\star$) achieves lower MSE than  \textit{LR} (\textit{FairLR}) as the former model considers the sample bias selection. More importantly, only $\textit{FairLR}^\star$ is able to achieve testing fairness given the fairness threshold.

\subsection{Performance Evaluation on Biased Ratio}

In this section, we evaluate how ratio $r =|\mathcal{D}_u|/|\mathcal{D}|$ would affect the performance of our $\textit{FairLR}^\star$ and baseline $\textit{FairLR}$ on the testing data. Note that larger $r$ indicates more bias in sample selection.  We conduct experiments on CRIME.  Figure  \ref{fig:curve} plots results of MD and MSED. In both plots, X-axis shows the varied $r$ values from 0.1 to 0.4, the left Y-axis shows MSE, and the right Y-axis shows the fairness metric (MD or MSED). Correspondingly, we use solid lines to represent MSE values and dashed lines to represent fairness values.
It is unsurprising to see that $\textit{FairLR}^\star$ always achieves better performance (smaller MSE and smaller MD or MSED) than \textit{FairLR}, as demonstrated in Figure \ref{fig:curve} that lines with symbol $\star$ locate below those with symbol $\diamond$. More importantly, for our $\textit{FairLR}^\star$, the fairness value (MD or MSED) and prediction error (MSE) are stable when $r$ increases, demonstrating the robustness of our $\textit{FairLR}^\star$ against sample selection bias. On the contrary, for \textit{FairLR}, both the unfairness and prediction error on the testing data increase when $r$ increases.

\section{Conclusions and Future Work}
In this paper, we have developed a framework for fair regression under sample selection bias when dependent variable values of a set of samples are missing. The framework adopts the classic Heckman model to correct sample selection bias and captures a variety of fairness notions via inequality and equality constraints. We applied the Lagrange duality theory to derive the dual convex optimization and showed the conditions of achieving strong duality for Pearson correlation.  For the two popular fairness notions, mean difference and mean squared error difference, we further derived explicit formulas without optimizing iteratively. Experimental results on three real-world datasets demonstrated our approach's effectiveness. In our future work, we will conduct theoretical analysis and empirical evaluation of density based fairness notions, e.g., SP and BGL, and notions for multiple sensitive attributes.
Some recent work \cite{grari2019fairness} proposed to use
Hirschfeld-Gebelein-Rényi Maximum (HGR) correlation coefficient as a regression fairness notion to  evaluate the independence between prediction and sensitive attributes. However, it is quite challenging to compute HGR. We can only get analytical solution for some certain distributions, e.g., jointly Gaussian distribution \cite{asoodeh2015maximal},  or apply approximation approaches. In our future work, we will study HGR in our framework. We will also study improved estimators~\cite{puhani2000heckman} that address the limitations of Heckman estimator, e.g., sensitivity of estimated coefficients with respect to the distributional assumptions on the error terms, and extend to nonlinear cases, e.g., kernel regression, in our fair regression.

\begin{ack}
This work was supported in part by NSF 1920920, 1939725, 1946391 and 2137335.
\end{ack}

\newpage

\appendix
\section{Fairness Metric}

\begin{definition}
\label{def:MD}
The mean difference (MD) of numeric prediction $\hat{y}$ in $\mathcal{D}$ by a binary protected attribute $a$ is defined as $\text{MD}(\hat{y}, a) = \mathbb{E}(\hat{y}|a = 0) - \mathbb{E}(\hat{y}|a = 1)$.
\end{definition}

\begin{definition}
\label{def:MSED}
The mean squared  error difference (MSED) of numeric prediction $\hat{y}$ in $\mathcal{D}$ by a binary protected attribute $a$ is defined as
$\text{MSED}(\hat{y}, a) =\mathbb{E}[(y - \hat{y})^2|a = 0] - \mathbb{E}[(y - \hat{y})^2|a = 1]$.
\end{definition}

\begin{definition}
\label{def:CC}
The correlation coefficient of numeric prediction $\hat{y}$ and numeric protected attribute $a$ is defined as $\rho_{\hat{y}a} = \frac{\mathbb{E}[(\hat{y} - \mu_{\hat{y}})(a - \mu_a)]}{\sigma_{\hat{y}} \sigma_{s}}$.
\end{definition}

\begin{definition}
\label{def:PCC}
The partial correlation coefficient of numeric prediction $\hat{y}$ and numeric protected attribute $a$ given $y$ is defined as $\rho_{\hat{y}a.y} = \frac{\rho_{\hat{y}a}-  \rho_{\hat{y}y} \rho_{ay}}{\sqrt{1-\rho^2_{\hat{y}{y}}}\sqrt{1-\rho^2_{ay}}}$.
\end{definition}

\begin{definition}
\label{def:SP}
The statistical parity (SP) is defined as $\text{SP} = \mathbb{P}[f(X) \ge z|A=a] - \mathbb{P}[f(X) \ge z] $ for all $a \in \mathcal{A}$ and $z \in [0,1]$.
\end{definition}

\begin{definition}
\label{def:BGL}
The bounded group loss (BGL) is defined as $ \text{BGL} = \mathbb{E}[l (f(X),Y)|A=a]$ for all $a \in \mathcal{A}$.
\end{definition}

\cite{DBLP:conf/icml/AgarwalDW19} presented two fairness definitions, statistical parity and bounded group loss. The statistical parity uses the departure of the CDF of $f(X)$ conditional on $A=a$ from the CDF of $f(X)$. When the departure is close to zero, the prediction is statistically independent of the protected attribute.  The bounded group loss which asks that the prediction error of any protected group stay below some pre-determined threshold.

\section{Conditional Expectation of Normal Distribution Variable}\label{sec: conditional_expectation}
The relationship between $u$ and $\epsilon$ is denoted as:
\begin{equation}
    (\epsilon, u) \sim N\{(\mu_{\epsilon}, \mu_u ), \begin{bmatrix}
\sigma_{\epsilon}^2 & \rho\sigma_u\sigma_{\epsilon}\\
\rho\sigma_u\sigma_{\epsilon} & \sigma_{u}^2
\end{bmatrix}\}
\end{equation}
Then we have the following:
\begin{equation}
\scalebox{0.75}{
$\begin{aligned}
&\mathbb{E}(\epsilon|u = \hat{u}) \\&=  \dfrac{\int_{-\infty}^{\infty} \epsilon \hspace{0.1cm}\text{exp}\Big( -\dfrac{1}{2(1-\rho^2)}\Big[ \dfrac{(\epsilon - \mu_{\epsilon})^2}{\sigma_{\epsilon}^2} + \dfrac{(\hat{u} - \mu_{u})^2}{\sigma_{u}^2} - \dfrac{2\rho(\epsilon - \mu_{\epsilon})(\hat{u} - \mu_u)}{\sigma_u^2}\Big] \Big)  \,d\epsilon }{\int_{-\infty}^{\infty} \hspace{0.1cm}\text{exp}\Big( -\dfrac{1}{2(1-\rho^2)}\Big[ \dfrac{(\epsilon - \mu_{\epsilon})^2}{\sigma_{\epsilon}^2} + \dfrac{(\hat{u} - \mu_{u})^2}{\sigma_{u}^2} - \dfrac{2\rho(\epsilon - \mu_{\epsilon})(\hat{u} - \mu_u)}{\sigma_u^2}\Big] \Big)  \,d\epsilon }
 \\
 &=\dfrac{\int_{-\infty}^{\infty} \epsilon \hspace{0.1cm}\text{exp}\Big( -\dfrac{1}{2(1-\rho^2)}\Big[ \dfrac{(\epsilon - \mu_{\epsilon})^2}{\sigma_{\epsilon}^2} - \dfrac{2\rho(\epsilon - \mu_{\epsilon})(\hat{u} - \mu_u)}{\sigma_u^2} + \dfrac{(\hat{u} - \mu_{u})^2}{\sigma_{u}^2} \Big] \Big)  \,d\epsilon }
 {\int_{-\infty}^{\infty} \hspace{0.1cm}\text{exp}\Big( -\dfrac{1}{2(1-\rho^2)}\Big[ \dfrac{(\epsilon - \mu_{\epsilon})^2}{\sigma_{\epsilon}^2} - \dfrac{2\rho(\epsilon - \mu_{\epsilon})(\hat{u} - \mu_u)}{\sigma_u^2} + \dfrac{(\hat{u} - \mu_{u})^2}{\sigma_{u}^2}  \Big] \Big)  \,d\epsilon } \\
 &=\dfrac{\int_{-\infty}^{\infty} \epsilon \hspace{0.1cm}\text{exp}\Big( -\dfrac{1}{2(1-\rho^2)}\Big[ \dfrac{(\epsilon - \mu_{\epsilon})}{\sigma_{\epsilon}} - \rho \dfrac{(\hat{u} - \mu_{u})}{\sigma_{u}} \Big]^2 \Big)  \,d\epsilon }
 {\int_{-\infty}^{\infty} \hspace{0.1cm}\text{exp}\Big( -\dfrac{1}{2(1-\rho^2)}\Big[ \dfrac{(\epsilon - \mu_{\epsilon})}{\sigma_{\epsilon}} - \rho \dfrac{(\hat{u} - \mu_{u})}{\sigma_{u}} \Big]^2 \Big)  \,d\epsilon }\\
 &= \dfrac{\int_{-\infty}^{\infty} \big(\epsilon + \mu_{\epsilon} + \sigma_{\epsilon}\rho \dfrac{(\hat{u} - \mu_{u})}{\sigma_{u}} \big) \hspace{0.1cm}\text{exp}\Big( -\dfrac{1}{2(1-\rho^2)}\dfrac{\epsilon^2}{\sigma_{\epsilon}^2} \Big)  \,d\epsilon }
 {\int_{-\infty}^{\infty} \hspace{0.1cm}\text{exp}\Big( -\dfrac{1}{2(1-\rho^2)}\dfrac{\epsilon^2}{\sigma_{\epsilon}^2} \Big)  \,d\epsilon}
\end{aligned}$}
\end{equation}
As
\begin{equation}
    \int_{-\infty}^{\infty} \epsilon \hspace{0.1cm}\text{exp}\Big( -\dfrac{1}{2(1-\rho^2)}\dfrac{\epsilon^2}{\sigma_{\epsilon}^2} \Big)  \,d\epsilon = 0
\end{equation}
Then we have the following:
\begin{equation}
    \mathbb{E}(\epsilon|u) = \mu_{\epsilon} + \rho\dfrac{\sigma_{\epsilon}}{\sigma_u}(u - \mu_u)
\end{equation}
Because we assume $u \sim N(0, 1)$, $\epsilon \sim N(0, \sigma_{\epsilon}^2)$, so we have
\begin{equation}
    \mathbb{E}(\epsilon|u) = \rho\sigma_{\epsilon}u
\end{equation}
With the law of iterated expectations we have:
\begin{equation}
\begin{aligned}
 \mathbb{E}(\epsilon|u > -\bm{x}_{1i}\gamma) &=  \mathbb{E}[\mathbb{E}(\epsilon|u)|u > -\bm{x}_{1i}\gamma)]\\
 &=\mathbb{E}[\rho\sigma_{\epsilon}u|u > -\bm{x}_{1i}\gamma)]\\
 &=\rho\sigma_{\epsilon}\mathbb{E}[u|u > -\bm{x}_{1i}\gamma)]\\
 &= \rho\sigma_{\epsilon} \dfrac{\phi(-\bm{x}_{1i}\bm{\gamma})}{1 - \Phi(-\bm{x}_{1i}\bm{\gamma})}\\
 &= \rho\sigma_{\epsilon} \dfrac{\phi(\bm{x}_{1i}\bm{\gamma})}{ \Phi(\bm{x}_{1i}\bm{\gamma})}
\end{aligned}
\end{equation}
where $\phi(\cdot)$ is the standard normal density function and $\Phi(\cdot)$ is the standard cumulative distribution function.

\section{Proofs of Closed Form for MD and MSED}
\begin{result}\label{result:MD}
For fair regression with the mean squared loss and $\text{MD}(\hat{y},a) =0$, we have the closed solution
\begin{equation}\label{eq:MD_Solution}
    \tilde{\bm{\beta}} = (\tilde{\bm{X}}_2^T\tilde{\bm{X}}_2)^{-1} (\tilde{\bm{X}}_2^T\bm{y}-\frac{\bm{d}^T(\tilde{\bm{X}}_2^T\tilde{\bm{X}}_2)^{-1} \tilde{\bm{X}}_2^T\bm{y}}{\bm{d}^T(\tilde{\bm{X}}_2^T\tilde{\bm{X}}_2)^{-1} \bm{d}} \bm{d})
\end{equation}
\end{result}
\begin{proof}
The fair Heckman prediction model can be described as:
\begin{equation}\label{eq:loss_under_fairness}
\begin{aligned}
&\min L(\tilde{\bm{\beta}}) = \sum_{i = 1}^{m}(\tilde{\bm{x}}_{2i}\tilde{\bm{\beta}}   - y_i)^2\\
&\text{subject to} \hspace{0.5cm} \dfrac{1}{m_0}{\sum_{i\in \mathcal{D}_0}\tilde{\bm{x}}_{2i}\tilde{\bm{\beta}}} = \dfrac{1}{m_1}{\sum_{i \in \mathcal{D}_1}\tilde{\bm{x}}_{2i}\tilde{\bm{\beta}}}
\end{aligned}
\end{equation}
where $m_0$ is the number of data in $\mathcal{D}_s$ with $a = 0$, $m_1$ is the number of data with $a = 1$, and $m = m_0 + m_1$.

We solve this optimization problem Eq. \ref{eq:loss_under_fairness} using Lagrange multipliers.
For convenience, we use $\bm{d}$ to denote
$\dfrac{1}{m_0}{\sum_{i\in \mathcal{D}_0}\tilde{\bm{x}}_{2i}} - \dfrac{1}{m_1}{\sum_{i \in \mathcal{D}_1}\tilde{\bm{x}}_{2i}}$. Then we can rewrite Eq. \ref{eq:loss_under_fairness} as the following constrained minimization problem:
\begin{equation}\label{eq: lagrange_loss}
    L(\tilde{\beta}) = \min \sum_{i = 1}^{m}(\tilde{\bm{\beta}}\tilde{\bm{x}}_{2i} - y_i)^2 + 2\lambda \bm{d}^T\tilde{\bm{\beta}}
\end{equation}
where $\lambda$ is the Lagrange multiplier.

By taking the partial derivatives of $j$th coefficient $\tilde{\beta}_j$ of $\tilde{\bm{\beta}}$:
\begin{equation}
\begin{aligned}
\dfrac{\partial L(\tilde{\bm{\beta}})}{\partial \tilde{\beta}_j}= \sum_{i = 1}^{m}2(\tilde{\bm{x}}_{2i}\tilde{\bm{\beta}} - y_i)\tilde{\bm{x}}_{2ij} + 2\lambda d_j
\end{aligned}
\end{equation}
where $\tilde{\bm{x}}_{2ij}$ is the $j$th component of $\tilde{\bm{x}}_{2i}$ and $d_j$ is the $j$th component of $\bm{d}$.
By setting the derivative to be zero for all $j$, we can get:
\begin{equation}\label{eq:gradient}
\begin{aligned}
 (\sum_{i = 1}^{m}\tilde{\bm{x}}_{2i}\tilde{\bm{x}}_{2ij})\tilde{\bm{\beta}} =\sum_{i = 1}^{m} y_i\tilde{\bm{x}}_{2ij} -\lambda d_j
\end{aligned}
\end{equation}
Thus we can rewrite Eq. \ref{eq:gradient} with matrix form:
\begin{equation}
    \tilde{\bm{X}}_2^T\tilde{\bm{X}}_2 \tilde{\bm{\beta}} = \tilde{\bm{X}}_2^T\bm{y}-\lambda \bm{d}
\end{equation}
where $\tilde{\bm{X}}_2$ is the matrix form of $\tilde{\bm{x}}_{2i}, i \in [m]$ and $\bm{y}$ is the vector form of $y_i, i \in [m]$. Therefore, we have:
\begin{equation}\label{eq:beta_lambda}
    \tilde{\bm{\beta}} = (\tilde{\bm{X}}_2^T\tilde{\bm{X}}_2)^{-1} (\tilde{\bm{X}}_2^T\bm{y}-\lambda \bm{d})
\end{equation}
We can also get solution of $\lambda$ using the fairness constraint $\bm{d}^T\tilde{\bm{\beta}}= 0$:
\begin{equation}\label{eq:lamda}
    \lambda = \frac{\bm{d}^T(\tilde{\bm{X}}_2^T\tilde{\bm{X}}_2)^{-1} \tilde{\bm{X}}_2^T\bm{y}}{\bm{d}^T(\tilde{\bm{X}}_2^T\tilde{\bm{X}}_2)^{-1} \bm{d}}
\end{equation}
By substituting $\lambda$ into Eq. \ref{eq:beta_lambda}, we have the closed solution.
\end{proof}

\begin{result}\label{result:MSED}
For fair regression with the mean squared loss and $\text{MSED}(\hat{y}, a) = 0$, we have the closed solution
\begin{equation}
\begin{aligned}
\tilde{\bm{\beta}} = &(\tilde{\bm{X}}_{2}^T\tilde{\bm{X}}_2 + \dfrac{\lambda}{m_0}(\tilde{\bm{X}}_{2}^0)^T\tilde{\bm{X}}_2^0 - \dfrac{\lambda}{m_1}(\tilde{\bm{X}}_{2}^1)^T\tilde{\bm{X}}_2^1)^{-1}(\tilde{\bm{X}}_{2}^T\bm{y} + \dfrac{\lambda}{m_0}(\tilde{\bm{X}}_{2}^0)^T\bm{y}_0 - \dfrac{\lambda}{m_1}(\tilde{\bm{X}}_{2}^1)^T\bm{y}_1)
\end{aligned}
\end{equation}
\end{result}
\begin{proof}
The fair Heckman prediction model can be described as:
\begin{equation}\label{eq:loss_under_equal_error}
\begin{aligned}
&\min L(\bm{\beta}) = \sum_{i = 1}^{m}(\tilde{\bm{x}}_{2i}\tilde{\bm{\beta}} - y_i)^2\\
&\text{subject to} \hspace{0.5cm} \dfrac{1}{m_0}{\sum_{i \in \mathcal{D}_0}(\tilde{\bm{x}}_{2i}\tilde{\bm{\beta}} - y_i)^2 } = \dfrac{1}{m_1}{\sum_{i\in \mathcal{D}_1}(\tilde{\bm{x}}_{2i}\tilde{\bm{\beta}} - y_i)^2}
\end{aligned}
\end{equation}
We  use the same notations as above and apply the  Lagrange multipliers:
\begin{equation}\label{eq: lagrange_loss_equal_error}
\begin{aligned}
L(\tilde{\bm{\beta}}) = &\min \sum_{i = 1}^{m}(\tilde{\bm{x}}_{2i}\tilde{\bm{\beta}} - y_i)^2 + \lambda (\dfrac{1}{m_0}{\sum_{i \in \mathcal{D}_0}(\tilde{\bm{x}}_{2i}\tilde{\bm{\beta}} - y_i)^2 } - \dfrac{1}{m_1}{\sum_{i\in \mathcal{D}_1}(\tilde{\bm{x}}_{2i}\tilde{\bm{\beta}} - y_i)^2})
\end{aligned}
\end{equation}
We can compute the derivatives of $\tilde{\bm{\beta}}$ with the matrix form and set it to be zero:
\begin{equation}
\begin{aligned}
2\tilde{\bm{X}}_{2}^T(\tilde{\bm{X}}_2\tilde{\bm{\beta}} - \bm{y}) + &\dfrac{2\lambda}{m_0}(
    \tilde{\bm{X}}_{2}^0)^T(\tilde{\bm{X}}_2^0\tilde{\bm{\beta}} -  \bm{y}_0) - \dfrac{2\lambda}{m_1}(\tilde{\bm{X}}_{2}^1)^T(\tilde{\bm{X}}_2^1\tilde{\bm{\beta}} -  \bm{y}_1) = 0
\end{aligned}
\end{equation}
where $\tilde{\bm{X}}_2^0$ is the matrix form of $\tilde{\bm{x}}_{2i}, i \in [m_0]$, $\tilde{\bm{X}}_2^1$ is the matrix form of $\tilde{\bm{x}}_{2i}, i \in [m_1]$, $\bm{y}_0$ is the vector form of $y_i, i \in [m_0]$, $\bm{y}_1$ is the vector form of $y_i, i \in [m_1]$, and $\bm{y}$ is the vector form of $y_i, i \in [m]$.
Then we can get:
\begin{equation}
\begin{aligned}
(\tilde{\bm{X}}_{2}^T\tilde{\bm{X}}_2 + &\dfrac{\lambda}{m_0}(\tilde{\bm{X}}_{2}^0)^T\tilde{\bm{X}}_2^0 - \dfrac{\lambda}{m_1}(\tilde{\bm{X}}_{2}^1)^T\tilde{\bm{X}}_2^1)\tilde{\bm{\beta}} = \tilde{\bm{X}}_{2}^T\bm{y} + \dfrac{\lambda}{m_0}(\tilde{\bm{X}}_{2}^0)^T\bm{y}_0 - \dfrac{\lambda}{m_1}(\tilde{\bm{X}}_{2}^1)^T\bm{y}_1
\end{aligned}
\end{equation}
Therefore, the solution of $\tilde{\bm{\beta}}$ is:
\begin{equation}
\begin{aligned}
\tilde{\bm{\beta}} = (\tilde{\bm{X}}_{2}^T\tilde{\bm{X}}_2 +& \dfrac{\lambda}{m_0}(\tilde{\bm{X}}_{2}^0)^T\tilde{\bm{X}}_2^0 - \dfrac{\lambda}{m_1}(\tilde{\bm{X}}_{2}^1)^T\tilde{\bm{X}}_2^1)^{-1}(\tilde{\bm{X}}_{2}^T\bm{y} + \dfrac{\lambda}{m_0}(\tilde{\bm{X}}_{2}^0)^T\bm{y}_0 - \dfrac{\lambda}{m_1}(\tilde{\bm{X}}_{2}^1)^T\bm{y}_1)
\end{aligned}
\end{equation}
\end{proof}

\section{Proof of Strong Duality}

\begin{result}
\label{result:slate}
For fair regression with the convex loss function and
the fairness inequality constraints (i.e., less than a user-specified threshold $\tau$), the strong duality holds for Pearson correlation if the linear relationship exists between $x$ and $a$.
\end{result}

\begin{proof}
Our proof is based on strong duality via Slater condition.
The Slater condition states that if a convex optimization problem has a feasible point $\tilde{\bm{\beta}}_0$ in the relative interior of the problem domain and every inequality constraint $g_i(\tilde{\bm{\beta}}) \le 0$ is strict at  $\tilde{\bm{\beta}}_0$, i.e.,   $g_i(\tilde{\bm{\beta}}_0) < 0$, then strong duality holds.

The correlation usually exists between $\tilde{\bm{x}}_2$ and $a$, and then the prediction based on $\tilde{\bm{x}}_2$ has disparate impact. We can remove the correlation between $\tilde{\bm{x}}_2$ and $a$ through the following regression:
\begin{equation}
    \hat{\bm{B}} = (\bm{A}^T\bm{A})^{-1}\tilde{\bm{X}}_{2}, \bm{U}=\tilde{\bm{X}}_{2} - \hat{\bm{B}}\bm{A}
\end{equation}
where $\bm{A} = (a_1, a_2, \cdots, a_n)$ and we define $\bm{u}_i$ as the $i$-th datapoint of $\bm{U}$. With the assumption that the linear relationship exists between $\tilde{\bm{x}}_2$ and $a$, it was proved by \cite{komiyama2018nonconvex} that ($\bm{u}$, $a$) has the same information with ($\tilde{\bm{x}}_2$, $a$) and the correlation between $\bm{u}$ and $a$ is $O(\dfrac{1}{\sqrt{n}})$.

Suppose the prediction outcome $\hat{y}$ is expressed as the following:
\begin{equation}
    \hat{y} = a\tilde{\bm{\beta}}_a + \bm{u}\tilde{\bm{\beta}}_u
\end{equation}
Then we can compute the Pearson coefficient as the following:
\begin{equation}\label{eq:pearson}
    \rho(\hat{y}, a) = \dfrac{Cov(\hat{y}, a)}{\sqrt{Var(\hat{y})Var(a)}}
\end{equation}
where $Cov(\hat{y}, a)$ is the correlation between $\hat{y}$ and $a$, $Var(\hat{y})$ is the variance of $\hat{y}$, and $Var(a)$ is the variance of $a$. $Cov(\hat{y}, a)$ is calculated as:
\begin{equation}
\begin{aligned}
Cov(\hat{y}, a) &= Cov(a\tilde{\bm{\beta}}_a + \bm{u}\tilde{\bm{\beta}}_u, a) \\
&=Cov(a\tilde{\bm{\beta}}_a, a) + Cov( \bm{u}\tilde{\bm{\beta}}_u, a)\\
&= \tilde{\bm{\beta}}_aVar(a) + 0\\
&=  \tilde{\bm{\beta}}_aVar(a)
\end{aligned}
\end{equation}
The variance of $\hat{y}$ is computed as:
\begin{equation}
\begin{aligned}
Var(\hat{y}) &= Var(a\tilde{\bm{\beta}}_a + \bm{u}\tilde{\bm{\beta}}_u) \\
&= Var(a\tilde{\bm{\beta}}_a) + Var(\bm{u}\tilde{\bm{\beta}}_u) \\&= \tilde{\bm{\beta}}_a^2Var(a) + \tilde{\bm{\beta}}_u^T \bm{V}_u\tilde{\bm{\beta}}_u
\end{aligned}
\end{equation}
where $\bm{V}_u$ is the covariances of $\bm{u}$.
Thus Eq. \ref{eq:pearson} can be written as:
\begin{equation}
\begin{aligned}
\rho(\hat{y}, a) &= \dfrac{Cov(\hat{y}, a)}{\sqrt{Var(\hat{y})Var(a)}}\\
&=\dfrac{\tilde{\bm{\beta}_a}Var(a) }{\sqrt{(\tilde{\bm{\beta}}_a^2Var(a) + \tilde{\bm{\beta}}_u^T \bm{V}_u\tilde{\bm{\beta}}_u)Var(a)}}\\
&=\dfrac{\tilde{\bm{\beta}_a}\sqrt{Var(a)} }{\sqrt{\tilde{\bm{\beta}}_a^2Var(a) + \tilde{\bm{\beta}}_u^T \bm{V}_u\tilde{\bm{\beta}}_u}}
\end{aligned}
\end{equation}
Up to now, we can write down the fairness regression subject to the fairness constraint of Pearson coefficient:
\begin{equation}\label{eq:fheckman}
\begin{aligned}
&\min_{\tilde{\bf{\beta}}}L(\tilde{\bm{\beta}}) = \sum_{i=1}^m{l[(f_h(\tilde{\bm{x}}_{2i}; \tilde{\bm{\beta}}), y)]}
\\
&\text{subject to} \hspace{0.2cm}\rho^2(\hat{y}, a) \leq \epsilon
\end{aligned}
\end{equation}
where $\epsilon$ is the threshold of the fairness metric. The fairness constraint $\rho^2(\hat{y}, a) \leq \epsilon$ is equivalent to:
\begin{equation}
    (1 - \epsilon)\tilde{\bm{\beta}}_a^2Var(a) - \epsilon \tilde{\bm{\beta}}_u^T \bm{V}_u\tilde{\bm{\beta}}_u \leq 0
\end{equation}

The Slater condition requires that $\{(\tilde{\bm{\beta}}_a, \tilde{\bm{\beta}}_u): (1 - \epsilon)\tilde{\bm{\beta}}_a^2Var(a)  - \epsilon \tilde{\bm{\beta}}_u^T \bm{V}_u\tilde{\bm{\beta}}_u < 0\} \neq  \emptyset$. It can be easily verified that Slater condition holds. For example, we can set $\tilde{\bm{\beta}}_a$ to be zero. Since $\bm{V}_u$ is symmetry and we can apply diagonal decomposition for
$\bm{V}_u$ and the eigenvalues of $\bm{V}_u$ cannot be all zero. Suppose the $j$th eigenvalue of $\bm{V}_u$ is non-zero, and we can set the corresponding $j$th component of $\tilde{\bm{\beta}}_u$ to be same sign with the $j$th eigenvalue, and set all other components of $\tilde{\bm{\beta}}_u$ to be zero, so that the  the Slater condition holds.
\end{proof}

\begin{table}[htb]
    \centering
    \caption{Attributes used for selection/prediction. Those with italic font are for prediction and those with either regular or italic font are for selection.}
    \scalebox{1.0}{
    \begin{tabular}{|p{0.15\linewidth} | p{0.75\linewidth}|}
     \hline
      \textbf{Dataset} & Attribute\\ \hline
      CRIME & population, householdsize, racepctblack,  racePctWhite,  racePctAsian,  \textit{racePctHisp},  \textit{agePct12t21},  \textit{agePct12t29},
      \textit{agePct16t24}\\
      & \textit{agePct65up},
         \textit{numbUrban},  \textit{pctUrban},  \textit{medIncome},  \textit{pctWWage},  \textit{pctWFarmSelf},  \textit{pctWInvInc},  \textit{pctWSocSec},  \textit{pctWPubAsst},  \textit{pctWRetire},  \textit{medFamInc} \\ \hline
      LAW &  cluster,  lsat,  ugpa,  zgpa,  \textit{fulltime},  \textit{fam\_inc},  \textit{age},  \textit{gender},  \textit{pass}  \\ \hline
         COMPAS &  decile\_score.1,  age\_cat\_25-45,  age\_cat\_45+, age\_cat\_25-,  c\_charge\_degree\_F,  c\_charge\_degree\_M, \textit{sex},  \textit{age},  \textit{juv\_fel\_count},  \textit{juv\_misd\_count},  \textit{juv\_other\_count},        \textit{priors\_count}, \textit{two\_year\_recid}     \\ \hline
     \end{tabular}}
    \label{tab:selection_prediction}
\end{table}

\section{Experiment}

Our experiments were carried out on the Dell PowerEdge C4130 with 2 Nvidia Tesla M10 GPU. Table \ref{tab:selection_prediction} shows information about attributes used in prediction/selection equations for three datasets.


\begin{thebibliography}{10}

\bibitem{Doe:2009:Online}
\url{http://archive.ics.uci.edu/ml/datasets/communities+and+crime}, 2009.

\bibitem{agarwal2018reductions}
Alekh Agarwal, Alina Beygelzimer, Miroslav Dud{\'\i}k, John Langford, and Hanna
  Wallach.
\newblock A reductions approach to fair classification.
\newblock In {\em International Conference on Machine Learning}, pages 60--69.
  PMLR, 2018.

\bibitem{DBLP:conf/icml/AgarwalDW19}
Alekh Agarwal, Miroslav Dud{\'{\i}}k, and Zhiwei~Steven Wu.
\newblock Fair regression: Quantitative definitions and reduction-based
  algorithms.
\newblock In {\em Proceedings of the 36th International Conference on Machine
  Learning {ICML}}, volume~97, pages 120--129. {PMLR}, 2019.

\bibitem{asoodeh2015maximal}
Shahab Asoodeh, Fady Alajaji, and Tam{\'a}s Linder.
\newblock On maximal correlation, mutual information and data privacy.
\newblock In {\em 2015 IEEE 14th Canadian workshop on information theory
  (CWIT)}, pages 27--31. IEEE, 2015.

\bibitem{berk2017convex}
Richard Berk, Hoda Heidari, Shahin Jabbari, Matthew Joseph, Michael Kearns,
  Jamie Morgenstern, Seth Neel, and Aaron Roth.
\newblock A convex framework for fair regression.
\newblock In {\em FAT ML}, 2018.

\bibitem{boyd2004convex}
Stephen Boyd, Stephen~P Boyd, and Lieven Vandenberghe.
\newblock {\em Convex optimization}.
\newblock Cambridge university press, 2004.

\bibitem{calders2013controlling}
Toon Calders, Asim Karim, Faisal Kamiran, Wasif Ali, and Xiangliang Zhang.
\newblock Controlling attribute effect in linear regression.
\newblock In {\em 2013 IEEE 13th international conference on data mining},
  pages 71--80. IEEE, 2013.

\bibitem{DBLP:conf/nips/ChzhenDHOP20a}
Evgenii Chzhen, Christophe Denis, Mohamed Hebiri, Luca Oneto, and Massimiliano
  Pontil.
\newblock Fair regression via plug-in estimator and recalibration with
  statistical guarantees.
\newblock In {\em Advances in Neural Information Processing Systems 33: Annual
  Conference on Neural Information Processing Systems 2020, NeurIPS 2020,
  December 6-12, 2020, virtual}, 2020.

\bibitem{chzhen2020fair}
Evgenii Chzhen, Christophe Denis, Mohamed Hebiri, Luca Oneto, and Massimiliano
  Pontil.
\newblock Fair regression with wasserstein barycenters.
\newblock {\em arXiv preprint arXiv:2006.07286}, 2020.

\bibitem{du2021robust}
Wei Du and Xintao Wu.
\newblock Fair and robust classification under sample selection bias.
\newblock In {\em Proceedings of the 2021 ACM International Conference on
  Information and Knowledge Management (CIKM)}. ACM, 2021.

\bibitem{du2021fairness}
Wei Du, Depeng Xu, Xintao Wu, and Hanghang Tong.
\newblock Fairness-aware agnostic federated learning.
\newblock In {\em Proceedings of the 2021 SIAM International Conference on Data
  Mining (SDM)}, pages 181--189. SIAM, 2021.

\bibitem{fitzsimons2019general}
Jack Fitzsimons, AbdulRahman Al~Ali, Michael Osborne, and Stephen Roberts.
\newblock A general framework for fair regression.
\newblock {\em Entropy}, 21(8):741, 2019.

\bibitem{grari2019fairness}
Vincent Grari, Boris Ruf, Sylvain Lamprier, and Marcin Detyniecki.
\newblock Fairness-aware neural rÃ©nyi minimization for continuous features.
\newblock In {\em IJCAI}, 2020.

\bibitem{hardt2016equality}
Moritz Hardt, Eric Price, and Nati Srebro.
\newblock Equality of opportunity in supervised learning.
\newblock In {\em Advances in neural information processing systems}, pages
  3315--3323, 2016.

\bibitem{heckman1979sample}
James~J Heckman.
\newblock Sample selection bias as a specification error.
\newblock {\em Econometrica: Journal of the econometric society}, pages
  153--161, 1979.

\bibitem{jiang2020identifying}
Heinrich Jiang and Ofir Nachum.
\newblock Identifying and correcting label bias in machine learning.
\newblock In {\em AISTATS}, 2020.

\bibitem{johnson2016impartial}
Kory~D Johnson, Dean~P Foster, and Robert~A Stine.
\newblock Impartial predictive modeling: Ensuring group fairness in arbitrary
  models.
\newblock {\em arXiv e-prints}, pages arXiv--1608, 2016.

\bibitem{kallus2018residual}
Nathan Kallus and Angela Zhou.
\newblock Residual unfairness in fair machine learning from prejudiced data.
\newblock In {\em ICML}, 2018.

\bibitem{kleinberg2016inherent}
Jon Kleinberg, Sendhil Mullainathan, and Manish Raghavan.
\newblock Inherent trade-offs in the fair determination of risk scores.
\newblock {\em arXiv preprint arXiv:1609.05807}, 2016.

\bibitem{komiyama2018nonconvex}
Junpei Komiyama, Akiko Takeda, Junya Honda, and Hajime Shimao.
\newblock Nonconvex optimization for regression with fairness constraints.
\newblock In {\em International conference on machine learning}, pages
  2737--2746. PMLR, 2018.

\bibitem{laforgue2019statistical}
Pierre Laforgue and Stephan Cl{\'e}men{\c{c}}on.
\newblock Statistical learning from biased training samples.
\newblock {\em arXiv preprint arXiv:1906.12304}, 2019.

\bibitem{compass}
J.~Larson, S.~Mattu, L.~Kirchner, and J.~Angwin.
\newblock Compas dataset.
\newblock \url{https://github.com/ propublica/compas-analysis}, 2017.

\bibitem{le2020computing}
Thibaut Le~Gouic and Jean-Michel Loubes.
\newblock Computing the price for fairness in a regression framework.
\newblock {\em arXiv preprint arXiv:2005.11720}, 2020.

\bibitem{mary2019fairness}
J{\'e}r{\'e}mie Mary, Cl{\'e}ment Calauzenes, and Noureddine El~Karoui.
\newblock Fairness-aware learning for continuous attributes and treatments.
\newblock In {\em International Conference on Machine Learning}, pages
  4382--4391. PMLR, 2019.

\bibitem{moreno2012unifying}
Jose~G Moreno-Torres, Troy Raeder, Roc{\'\i}o Alaiz-Rodr{\'\i}guez, Nitesh~V
  Chawla, and Francisco Herrera.
\newblock A unifying view on dataset shift in classification.
\newblock {\em Pattern recognition}, 45(1):521--530, 2012.

\bibitem{narasimhan2020pairwise}
Harikrishna Narasimhan, Andrew Cotter, Maya Gupta, and Serena Wang.
\newblock Pairwise fairness for ranking and regression.
\newblock In {\em Proceedings of the AAAI Conference on Artificial
  Intelligence}, 2020.

\bibitem{puhani2000heckman}
Patrick Puhani.
\newblock The heckman correction for sample selection and its critique.
\newblock {\em Journal of economic surveys}, 14(1):53--68, 2000.

\bibitem{DBLP:conf/aaai/RezaeiFMZ20}
Ashkan Rezaei, Rizal Fathony, Omid Memarrast, and Brian~D. Ziebart.
\newblock Fairness for robust log loss classification.
\newblock In {\em AAAI}, 2020.

\bibitem{schumann2019transfer}
Candice Schumann, Xuezhi Wang, Alex Beutel, Jilin Chen, Hai Qian, and Ed~H Chi.
\newblock Transfer of machine learning fairness across domains.
\newblock {\em arXiv preprint arXiv:1906.09688}, 2019.

\bibitem{DBLP:journals/corr/abs-2002-06200}
Daniel Steinberg, Alistair Reid, Simon O'Callaghan, Finnian Lattimore, Lachlan
  McCalman, and Tib{\'{e}}rio~S. Caetano.
\newblock Fast fair regression via efficient approximations of mutual
  information.
\newblock {\em CoRR}, abs/2002.06200, 2020.

\bibitem{taskesen2020distributionally}
Bahar Taskesen, Viet~Anh Nguyen, Daniel Kuhn, and Jose Blanchet.
\newblock A distributionally robust approach to fair classification.
\newblock {\em arXiv preprint arXiv:2007.09530}, 2020.

\bibitem{wen2014robust}
Junfeng Wen, Chun-Nam Yu, and Russell Greiner.
\newblock Robust learning under uncertain test distributions: Relating
  covariate shift to model misspecification.
\newblock In {\em ICML}, 2014.

\bibitem{wightman1998lsac}
Linda~F Wightman.
\newblock Lsac national longitudinal bar passage study. lsac research report
  series.
\newblock 1998.

\bibitem{zhao2020unfairness}
Chen Zhao and Feng Chen.
\newblock Unfairness discovery and prevention for few-shot regression.
\newblock In {\em 2020 IEEE International Conference on Knowledge Graph
  (ICKG)}, pages 137--144. IEEE, 2020.

\end{thebibliography}

\end{document}